\documentclass[final,5p,times,twocolumn]{elsarticle} 
\usepackage{amssymb}
\usepackage{lipsum}
\usepackage{soul}
\usepackage{comment}
\usepackage{dblfloatfix}
\usepackage{hyperref}

\usepackage[utf8]{inputenc}

\usepackage{amsmath}

\usepackage{amsfonts}

\usepackage{amssymb}

\usepackage{graphicx}

\usepackage{booktabs}

\usepackage{xcolor}

\usepackage{array}

\usepackage{calc,ragged2e}

\usepackage{subcaption}

\usepackage{float}

\usepackage{tabu}

\usepackage{etoolbox}
\usepackage{multirow}
\usepackage{color}
\usepackage{makecell}
\usepackage{comment}

\definecolor{gris1}{gray}{0}

\definecolor{gris2}{gray}{0.45}

\definecolor{gris3}{gray}{0.6}

\definecolor{gris4}{gray}{0.75}

\definecolor{gris5}{gray}{0.9}

\makeatletter

\patchcmd{\@makecaption}

{\scshape}

{}

{}

{}

\journal{Journal of Visual Communication and Image Representation}

\begin{document}

\begin{frontmatter}

\title{Evaluating ADC-only deep learning pipelines for breast cancer detection and segmentation using standalone diffusion-weighted MRI}

\author[uniovi_cs,uniovi_bme]{Pablo García Marcos}
\ead{garciamarpablo@uniovi.es}

\author[uniovi_ee]{Paula Puerta González}
\ead{puertapaula@uniovi.es}

\author[unicoru_ma,unitx]{Guillermo Lorenzo}
\ead{guillermo.lorenzo@udc.es}

\author[purdue_me,purdue_bme,purdue_ccr]{Hector Gomez}
\ead{hectorgomez@purdue.edu}

\author[huca_ra]{Covadonga del Camino}
\ead{caminocovadonga@uniovi.es}

\author[uniovi_cs,uniovi_bme]{Angel Rio-Alvarez \corref{cor1}}
\ead{rioangel@uniovi.es}

\author[uniovi_ee,uniovi_bme]{Víctor M. González}
\ead{vmsuarez@uniovi.es}

\cortext[cor1]{Corresponding author}

\address[uniovi_cs]{Computer Sciences Department, University of Oviedo, Asturias,  Spain}
\address[uniovi_ee]{Electrical Engineering Department, University of Oviedo, Asturias,  Spain}
\address[uniovi_bme]{Biomedical Engineering Center (BME), University of Oviedo, Asturias, Spain}
\address[unicoru_ma]{Group of Numerical Methods in Engineering, Department of Mathematics, University of A Coruña, A Coruña, Spain}
\address[unitx]{Oden Institute for Computational Engineering and Sciences, The University of Texas at Austin, Austin, Texas, USA}
\address[purdue_me]{School of Mechanical Engineering, Purdue University, West Lafayette, Indiana, USA}
\address[purdue_bme]{Weldon School of Biomedical Engineering, Purdue University, West Lafayette, Indiana, USA}
\address[purdue_ccr]{Purdue Center for Cancer Research, Purdue University, West Lafayette, Indiana, USA}
\address[huca_ra] {Radiodiagnostic Service. Asturias Central University Hospital (HUCA). Oviedo, Spain}
 
\begin{abstract}
\setlength{\parindent}{0pt}

\textbf{Background and Objectives}
Dynamic contrast-enhanced (DCE) imaging is the gold standard technique for the detection and characterization of breast cancer using magnetic resonance imaging (MRI). However, DCE-MRI requires long acquisition times and the administration of contrast into the bloodstream, which can cause allergic reactions. Alternatively, diffusion-weighted MRI (DW-MRI) is a standard complementary technique for breast MRI that does not require contrast, has shorter acquisition times, and enables calculation of apparent diffusion coefficient (ADC) maps that correlate with tumor cellularity. Yet, despite these technical advantages, deep learning research has focused on DCE-based models and has barely explored the tumor detection performance of DW-MRI and ADC maps either in combination with DCE-MRI or as standalone alternatives. Here, we evaluate the application of different state-of-the-art deep learning techniques for detection and segmentation of breast cancer using ADC-only images. This is, to our knowledge, the first comprehensive evaluation of ADC-only breast cancer pipelines for classification, detection, and segmentation tasks.

\textbf{Methods}
The evaluation was conducted using various well-established computer vision algorithms, including YOLOv8 for detection and nnU-Net for segmentation, for the classification, detection and segmentation of breast cancer. The test cohort is created from the public dataset ACRIN 6698 ($n=385$).

\textbf{Results}
Our results show that slice-based classification with the object detection model achieves the best performance in tumor detection followed by tumor localization using the same model, with mean precisions of 0.94 and 0.88. Additionally, 2D and 3D convolutional networks for tumor segmentation exhibited Dice coefficients of, at most, 0.69, which underperform compared to analogous DCE-based models in the literature. 

\textbf{Conclusions}
Diffusion weighted imaging may be used as a tool for contrast-free pipelines aimed at detection and localization of breast cancer. Meanwhile, the limited resolution and anisotropy of ADC acquisitions remain major challenges for robust volumetric segmentation.
\begin{comment}
Short Abstract

Dynamic contrast-enhanced (DCE) MRI is the reference technique for detecting and characterizing breast cancer, but it requires long acquisitions and contrast injection, which may cause adverse reactions. Diffusion-weighted MRI (DW-MRI) is routinely acquired, faster, and does not require contrast, while providing apparent diffusion coefficient (ADC) maps that reflect tumor cellularity. However, despite these advantages, deep learning efforts have focused almost exclusively on DCE-based models, and the ability of DW-MRI and ADC maps to detect tumors—either alone or combined with DCE-MRI—remains insufficiently explored. We introduce deep learning models trained solely on ADC maps for the automatic detection and segmentation of breast cancer, using 2D and 3D convolutional architectures and object detection methods. Experiments on the ACRIN 6698 dataset (n = 385) show that ADC-based object detection achieves high tumor detection and localization performance. Although segmentation remains limited due to the intrinsic properties of ADC imaging, our findings demonstrate the potential of ADC-only approaches as a non-contrast alternative for breast cancer detection.
\end{comment}
\end{abstract}


\end{frontmatter}

%\tableofcontents

%% \linenumbers

%% main text

\section{Introduction}\label{sec: intro}

Breast cancer is the most diagnosed tumor type and the leading cause of cancer-related death among women worldwide. For example, in 2022 alone, there were around 2,300,000 new cases along with 665,000 related deaths, respectively representing 23.8\%  and 15.4\% of all new cancer diagnoses and deaths among women around the globe \cite{bray_global_2024}.
Breast tumors are characterized by the uncontrolled proliferation of malignant cells within breast tissue, which can form tumors capable of invading surrounding structures and metastasizing distant organs. 
The development and progression of breast cancer are often influenced by a combination of genetic, hormonal, and environmental factors. Additionally, breast cancer is a highly heterogeneous disease, comprising multiple subtypes with distinct genetic and molecular profiles, which play a critical role in determining treatment strategies and prognosis \cite{desantis_breast_2019}.

Early detection and continuous monitoring are essential for the effective clinical management of breast cancer, with magnetic resonance imaging (MRI) playing a central role in both diagnosis and treatment. 
In particular, dynamic contrast-enhanced MRI (DCE-MRI) is considered the gold standard for breast cancer detection and characterization. This MRI technique improves the distinction between benign and malignant tissue by enhancing the visualization of blood flow patterns, which are often irregular in malignant tumors \cite{mann_breast_2008, washington_role_2024}. 
However, DCE-MRI also presents certain drawbacks, including the requirement for contrast agents (which may trigger allergic reactions in some patients), high operational costs, and long acquisition times due to the need for multiple time-point acquisitions followed by complex post-processing \cite{hashem_can_2021}.
In addition to DCE-MRI, diffusion-weighted MRI (DW-MRI) is a standard imaging technique for breast MRI that relies on the ability of water molecules to move within tissues under gradients of the magnetic field \cite{padhani_diffusion-weighted_2010}. 
This phenomenon can also be measured through apparent diffusion coefficient (ADC) maps, which are routinely calculated from DW-MRI data.
Since tumor tissues typically exhibit restricted water diffusion, they are enhanced in DW-MRI and exhibit low ADC.
Moreover, ADC values have been correlated to tumor cellularity and other tissue properties of interest for the breast cancer diagnosis \cite{padhani_diffusion-weighted_2010, horvat_diffusionweighted_2019}.
Hence, DW-MRI offers a faster, non-contrast, alternative to detect and characterize breast tumors, especially for patients for whom contrast agent administration is contraindicated \cite{abdelgawad_can_2018}. 

In recent years, applications of deep learning (DL) to biomedical imaging and disease diagnosis have experienced rapid growth, fueled by advances in computing power and the availability of large-scale datasets for training \cite{cox_neural_2014,voulodimos_deep_2018,litjens_survey_2017,wu2025critical}.
In the context of breast cancer, DL-based techniques are increasingly applied to assist in tumor detection, segmentation, diagnosis, and prognosis \cite{bi2019artificial,jones2022applying,wu2025critical}. 
Given its standardized diagnostic role, DCE-MRI is the main MRI sequence for developing DL tools for automatic detection and characterization of breast cancer, while there is a dearth of efforts that aim to construct accurate DL-based models for DW-MRI and ADC maps either alone or combined with DCE-MRI data \cite{bi2019artificial,jones2022applying,zhu_development_2022}.
Given the advantages of DW-MRI over DCE-MRI (particularly its safety, speed, and contrast-agent-free imaging) the application of DL techniques on DW-MRI data holds significant potential to enhance the detection and segmentation of breast cancer and cancer-affected tissue. Still, the limitations of ADC-only DL remain relatively unexplored and could be further documented.

Thus, the objective of this study is to explore and evaluate the viability of various DL approaches aimed at breast cancer identification using ADC imaging. These approaches range from object detection (OD) \cite{litjens_survey_2017, zaidi_survey_2022} for identifying and localizing the presence of cancer to convolutional neural networks (CNNs) for semantic segmentation of tumors \cite{guo_review_2018}.

The rest of this paper is structured as follows. Section~\ref{sec: background} reviews state-of-the-art computer vision techniques in medical imaging. Section~\ref{sec: matdata} describes the dataset used in this work. Section~\ref{sec: dlmethods} presents our DL models, including their architecture and implementation details and defines the metrics used to define the performance of our DL-based models. In Section~\ref{sec: results}, we present and discuss the results of our study. Finally, Section~\ref{sec: conclusion} provides final conclusions and future avenues of research.

\section{Related work}\label{sec: background}

Automated lesion detection and localization play a critical role in clinical workflows for breast cancer diagnosis, and OD techniques have found widespread use for these specific tasks.
For example, Frank \emph{et al.} \cite{frank_deep_2023} explored the integration of OD and slice classification by combining a CNN with a dedicated detection algorithm. They applied their approach to mammographic images, showcasing strong performance in accurately localizing breast mass lesions while reducing false positives.
Similarly, Kebede \emph{et al.} \cite{kebede_dual_2024} proposed a dual-view DL model specifically designed for mammography-based breast cancer detection. This model was based on a combination of EfficientNet ensemble classifiers and the YOLOv5 detection framework, leveraging craniocaudal and mediolateral oblique views to enhance the system’s ability to localize and classify suspicious regions. 
Additionally, Prinzi \emph{et al.} \cite{prinzi_yolo-based_2024} employed a YOLO-based architecture combined with transfer learning strategies to automate the detection of breast cancer lesions in mammograms. This approach displayed a significantly boost detection precision thanks to fine-tuning OD models. 
Using digital breast tomosynthesis,  Alashban \emph{et al.} \cite{alashban_breast_2024} introduced a two-stage deep learning pipeline, which combined a modified VGG19 classifier with an optimized YOLOv5 detection module augmented by a convolutional block attention module (CBAM). 
Thus, while there are many efforts in developing OD models for breast cancer detection and localization based on mammographies and digital breast tomosynthesis, there is a dearth of OD approaches leveraging DW-MRIs or ADC maps.

Beyond OD, existing computer vision approaches for the detection and segmentation of tumors on breast MRI have traditionally relied on diverse machine learning methods applied to DCE-MRI data. 
For instance, Militello \emph{et al.} \cite{militello_unsupervised_2021} explored four unsupervised learning methods to segment breast tumors on DCE-MRI data, including split-and-merge combined with region growing, k-means, as well as classical and spatial fuzzy c-means.
Additionally, Ma \emph{et al.} \cite{ma_radiomics_2021} employed a 3D U-Net to delineate the tumor boundaries on post-contrast DCE images in order to identify triple-negative and non-triple-negative breast cancers. 
Other advanced architectures for automatic breast cancer segmentation on DCE-MRI data include transformer-based models for global context extraction combined with CNN-based feature learning \cite{qin_joint_2022} and models that integrate spatial and temporal information to improve segmentation accuracy (e.g., pre and post-contrast DCE images) \cite{huang_joint-phase_2023}. 

Although DCE-MRI remains the most common imaging modality for research on automatic detection and segmentation of tumors on breast MRI data, some studies have explored hybrid approaches combining DCE-MRI and DW-MRI to improve performance.
For example, Cai \emph{et al.} \cite{cai_diagnosis_2014} investigated standard machine learning models to classify breast lesions on DCE-MRI and DW-MRI data, including support vector machines, Na\"ive Bayes, k-nearest neighbors, and logistic regression. 
To this end, the authors previously used a hybrid filter-wrapper method to select relevant features from DCE-MRI and DW-MRI data, such as ADC, early phase enhancement, signal enhancement ratio, as well as morphological and texture features from the lesion segmented on DCE images using a combination of fuzzy c-means and gradient vector flow snake algorithm.  
Furthermore, Zhu \emph{et al.} \cite{zhu_development_2022} segmented lesions on breast MRI on both DCE-MRI and DW-MRI by leveraging a 2D V-Net framework and an Attention U-Net model, respectively. 
The authors then used a 3D version of ResNet informed by the segmentations obtained from DCE-MRI and DW-MRI data to classify breast MRI lesions as benign or malignant.
Additionally, Wang \emph{et al.} \cite{wang_breast_2024} proposed a hybrid DL network capable of leveraging spatial-temporal features from DCE-MRI and DW-MRI for breast tumor segmentation.
This network was composed of three interconnected modules: a multi-sequence encoder, a multi-scale feature embedding module, and a decoder that produced the final breast tumor segmentation.

However, despite the advances in DL-based automatic tools for identifying and delineating breast tumors in DCE-MRI data alone or combined with DW-MRI, research on DL models exclusively using DW-MRI data for these tasks remains limited. Recent studies have demonstrated the effectiveness of attention-based and multimodal deep learning architectures in breast ultrasound \cite{nissar_swineff-attentionnet_2026}, mammography \cite{nissar_mob-cbam_2024}, and thermal imaging \cite{nissar2024lcscs}, while recent surveys have highlighted these directions as major trends in breast cancer image analysis \cite{nissar2022recenttrends}. Nevertheless, these developments have not yet translated into a substantial body of work focused on lesion detection and localization directly from DW-MRI or ADC maps. Some studies \cite{kapsner_image_2023, olthof_optimizing_2024} have applied DL methods to DW-MRI in the context of breast cancer diagnosis, but their focus lies primarily in image pre-processing and curation rather than tumor detection and segmentation.

\section{Materials and Data}\label{sec: matdata}

\begin{figure}[h]
	\centering
	\includegraphics[width=1\linewidth]{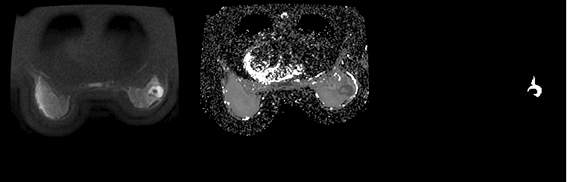}
	\caption{ Slice of a study of the dataset in DW (left), ADC (center) and mask (right) formats.}
	\label{fig:dataset}
\end{figure}

We employed the ACRIN 6698 dataset, which is publicly available at The Cancer Imaging Archive~\cite{noauthor_acrin_nodate}.
This dataset was created as part of a multicenter clinical trial for neoadjuvant chemotherapy of breast cancer to evaluate the reliability of ADC measurements and their capability to predict therapeutic response during the course of treatment \cite{newitt_testretest_2019, partridge_diffusion-weighted_2018}.
The dataset includes imaging studies from $n=385$ breast cancer patients who underwent up to four MRI examinations at key clinical timepoints: prior to treatment (T0), early in treatment after three cycles of paclitaxel (T1), mid-treatment between paclitaxel and doxorubicin-cyclophosphamide regimens (T2), and after completing the full treatment course (T3). 
The data from each imaging session was collected using several types of MRI sequences, including the widely used DCE-MRI and DW-MRI.
Each patient dataset contains the original DW-MRI scans, the corresponding ADC maps calculated with the classic monoexponential decay model, and tumor segmentation masks for DW-MRI and ADC maps created manually by experts.
These tumor segmentations consist of a binary mask, where tumor regions are labeled as the positive class and all other tissues as the negative class.
Figure~\ref{fig:dataset} shows an slice of DW-MRI data, ADC map, and tumor segmentation for a patient. 
Details on MRI acquisition have been provided in previous works \cite{newitt_testretest_2019, partridge_diffusion-weighted_2018}.

Thus, ACRIN 6698 offers all the information needed for an evaluation of ADC-only DL techniques and applications. Still, some limitations should be mentioned. First, ACRIN 6698 is a multicenter and highly heterogeneous study, which may be result in performance degradation due to scanner variability and acquisition heterogeneity. Nevertheless, these factors lend greater significance to the results, as the performance should be closer to a real life multi-center application than those obtained with more homogeneous studies. Furthermore, the anisotropy in the dataset, mainly characterized by a large inter-slice gap of 4 mm, may significantly degrade performance, especially when applying 3D semantic segmentation techniques.

In this study we used the data available for the pre-treatment scan (T0), which was prepared for model training through a pre-processing pipeline. 
First, to ensure the reliability of the data, a filtering process was performed, removing the images whose segmentation mask was not labeled as "high quality" in the database.
The images were then resized, normalized, and matched with their corresponding segmentation masks. The images were resized to a shape of 256$\times$256 and normalized to the interval $\left [0,1\right ]$ using a contrast stretch, where the top 0.2\% of values were flattened to 1.0 to remove imaging artifacts \cite{kamnitsas_efficient_2017, zhao_deep_2018}. Additionally, 3D volumes were created from the original 2D ADC maps, DW-MRI data, and segmentation masks to evaluate the performance of both 2D and 3D semantic segmentation. For segmentation, both the 2D images and the 3D volumes were used at their full 256×256 resolution, without applying any region-of-interest cropping, so that no ground-truth information is required to localize the tumor at inference time. 
For OD, bounding box annotations enclosing the tumor are required to provide spatial context for tumor detection and localization tasks.
These annotations were generated as part of the pre-processing pipeline by finding the minimal box capable of enclosing the tumor defined in the mask and applying a margin of 6 voxels.

\section{Methods}\label{sec: dlmethods}

This section describes our deep learning algorithms for slice classification, tumor detection, and tumor segmentation. In all cases, the input image(s) is (are) ADC maps with one single $b$-value.

\subsection{Deep learning architectures}

\subsubsection{Architecture for slice classification and object detection}

We use YOLO for slice classification and object detection. YOLO is a fully convolutional network designed for real-time object detection. Unlike traditional architectures for object detection, YOLO directly predicts bounding box coordinates and class probabilities for an entire image in a single forward pass \cite{Redmon2016,flores-calero_traffic_2024}. The YOLO architecture divides the input image into a grid of cells. Each cell is responsible for predicting a predefined number of bounding boxes. In the output of the model, the bounding boxes are accompanied by class probabilities, which allow the model to classify the detected objects and confidence scores. The confidence scores represent the probability that the bounding box contains an object and the accuracy of the bounding box’s spatial alignment relative to the object. We tested multiple versions and model sizes \cite{hussain_yolo-v1_2023,sirisha_statistical_2023}. We found that the large model of YOLOv8 led to the best performance in our tests.

To perform slice classification we proceed as follows: We run YOLO using single slices as input. The model outputs one or several bounding boxes with their corresponding probabilities for each slice. If the none of those probabilities is higher than 0.25, the slice is classified as tumor free. If one or more of the probabilities is higher than 0.25, the slice is classified as featuring a tumor.

\subsubsection{Architecture for semantic segmentation}

We use the U-Net architecture for semantic segmentation. U-Net is a convolutional network for biomedical image segmentation. The architecture is named for its distinctive U shape, which consists of an encoder and a decoder \cite{Ronneberger2015}. The encoder comprises convolutional layers and down-sampling operations, which progressively reduce the spatial dimensions of the input image and transform it into abstract feature representations. The decoder reconstructs the spatial resolution of the input image through up-sampling techniques. The outcome is a pixel-level segmentation map, where each pixel is assigned a class label.

We evaluated several variations of the U-Net architecture. As an initial approach, we adapted the original U-Net design \cite{agrawal_segmentation_2022} for both 2D and 3D inputs, operating on tumor-centered sub-volumes obtained via ground-truth-guided cropping. Given the poor performance achieved with this approach, and to avoid relying on ground-truth information at inference time, nnU-Net was subsequently employed on the full-resolution images and volumes described in Section \ref{sec: matdata}. The nnU-Net framework provides a self-configuring approach to semantic segmentation based on U-Net and is renowned for its state-of-the-art performance across diverse benchmarks \cite{isensee2021nnunet}. nnU-Net was included to evaluate segmentation performance due to its prominence as a baseline method for both 2D and 3D biomedical image segmentation.

\subsection{Training, validation and test data. Optimization algorithms}

We perform a $k$-fold cross-validation with $k=5$. For each fold we used 20\% of the patients as test. The remaining (non-test) patients were split into 3/4 for training and 1/4 for validation during training. 

YOLO was trained using a stochastic gradient descent (SGD) optimizer with an initial learning rate of 0.01. The model was trained for 200 epochs.

%The custom U-Net models were trained using the Adam optimizer was used with an initial learning rate of $10^{-4}$. We evaluated multiple loss functions including Dice loss, focal loss, binary cross-entropy (BCE) with logits loss, and combo loss, where combo loss refers to a combination of BCE with logits loss and Dice loss. In the BCE loss, a class weight of 10 was applied to the positive class to address class imbalance. We found that a combo loss weighted at 0.6 for BCE and 0.4 for Dice loss performed best. The 3D model was trained for 70 epochs, while the 2D model was trained for 50 epochs. 

The nnUnet model was trained using its default configuration settings "3d\_fullres" and "2d" for version 2. For the nnUnet 20\% of the dataset was used for validation in each fold, with the performance being calculated as the mean best epoch validation performance across the five folds.

\subsection{Model performance metrics}

We will evaluate our models by comparing them with against ground truth annotations. Our evaluation metrics include
\begin{equation}
	\text{Recall} = \frac{\text{TP}}{\text{TP + FN}}
	\label{eq:recall}
\end{equation}
\begin{equation}
	\text{Precision} = \frac{\text{TP}}{\text{TP + FP}}
	\label{eq:precision}
\end{equation}
\begin{equation}
	\text{Accuracy} = \frac{\text{TP + TN}}{\text{TP + TN + FP + FN}}
	\label{eq:accuracy}
\end{equation}
\begin{equation}
	\text{Specificity} = \frac{\text{TN}}{\text{TN + FP}}
	\label{eq:specificity}
\end{equation}
where TP, TN, FP and FN, indicate, respectively, true positive, true negative, false positive and false negative. For classification, the outcome classes (TP, TN, FP and FN) are defined at the slice level. For object detection, the outcome classes are defined at the bounding box level: A TP corresponds to a predicted bounding box that matches the ground truth with an overlap larger than 50\%. TNs are not relevant in object detection. A FP occurs when a predicted box does not correspond to any ground truth box with sufficient overlap. A FN occurs when a ground truth object is not detected by the model. In semantic segmentation, we define the outcome classes at the voxel level. To define, TP, FP and FN, we consider only boxes that had a confidence score of 0.25 or higher. All other boxes are disregarded and do not contribute to the metrics. 

To evaluate the performance of our object detection model, in addition to the metrics defined in Eqs.~\eqref{eq:recall}--\eqref{eq:precision}, we also use the average precision (AP). To compute AP, we determine precision and recall at all values of the confidence threshold. This defines a precision-recall curve, and the AP is the area under that curve.

%This last metric is calculated by first determining precision and recall at any intersection over union (IoU) thresholds, typically ranging from 0.5 to 0.95. The average precision (AP) is computed as the area under the precision-recall curve, which is then averaged across different recall levels \ref{eq:ap}. The mean of these AP values across all classes gives the mAP score \ref{eq:map}. For this research, we adopted an IoU threshold of 0.5 (mAP@0.5), where predictions with an Intersection over Union of at least 50\% with the ground truth are considered correct.

%\begin{equation}
%	AP = \int_0^1 p(r) \, dr \\
%	\label{eq:ap}
%\end{equation}
%
%\begin{equation}
	%mAP = \frac{1}{C} \sum_{i=1}^{C} AP_i
	%\label{eq:map}
%\end{equation}

To evaluate the performance of our semantic segmentation model, we use 
\begin{equation}
	\text{Dice Coefficient} = \frac{2|A \cap B|}{|A| + |B|}
	\label{eq:dice}
\end{equation}
where $A$ represents the set of voxels predicted as cancerous by our model and $B$ denotes the set of voxels labeled as cancerous in the ground truth \cite{xia_early_2024,muller_towards_2022}.

\begin{table}
	\centering
	\caption{Performance of the model in slice classification.}
	\begin{tabular}{|c|c|c|c|c|}
		\hline
		\textbf{Fold} & \textbf{Precision} & \textbf{Recall} & \textbf{Accuracy} & \textbf{Specificity}\\
		\hline
		1 & 0.945 & 0.896 & 0.933 & 0.960 \\
		2 & 0.958 & 0.865 & 0.921 & 0.969\\
		3 & 0.954 & 0.823 & 0.910 & 0.971\\
		4 & 0.916 & 0.813 & 0.881 & 0.938\\
		5 & 0.939 & 0.826 & 0.905 & 0.962\\
		\textbf{Mean} & \textbf{0.942} & \textbf{0.844} & \textbf{0.910} & \textbf{0.960} \\
		\hline
	\end{tabular}
	\label{tab:classification}
\end{table}

\section{Results}\label{sec: results}

Here, we report the results of our models for slice classification, tumor detection, and semantic segmentation. %These experiments are designed to evaluate the usefulness of DW-MRI and ADC maps for different levels of breast cancer identification and analysis using deep learning techniques. The order of presentation follows a logical progression from simpler tasks, such as classifying individual slices, to more complex tasks like full tumor segmentation. After summarizing the contribution of each experiment, the relevant performance metrics are shown and discussed.

\subsection{Slice classification}
We investigate the ability of the model to classify individual ADC slices as healthy or cancerous. This task is simpler than tumor detection or segmentation but offers useful insight into the potential of DW-MRI for preliminary screening.

The model shows very strong performance, achieving a mean precision of 94.2\% and an overall accuracy of 91.0\%; see Table~\ref{tab:classification}. These values support the idea that ADC maps may be used as a non-invasive screening tool, particularly when contrast-based imaging is not recommended.

\begin{figure}
    \centering
    \begin{subfigure}[b]{0.45\linewidth}
        \centering
        \includegraphics[width=\linewidth]{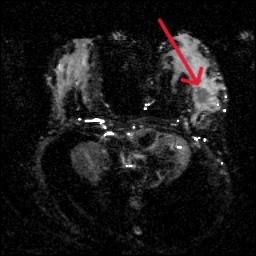}
        \caption{True positive prediction.}
        \label{fig:tp}
    \end{subfigure}
    \hfill
    \begin{subfigure}[b]{0.45\linewidth}
        \centering
        \includegraphics[width=\linewidth]{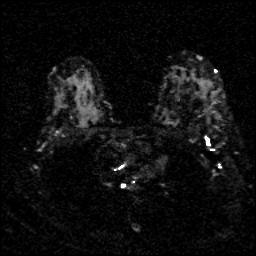}
        \caption{True negative prediction.}
        \label{fig:tn}
    \end{subfigure}
    \caption{Classification results: ADC slices correctly classified by the model.}
    \label{fig:class_success}
\end{figure}
\begin{figure}
    \centering
    \begin{subfigure}[b]{0.45\linewidth}
        \centering
        \includegraphics[width=\linewidth]{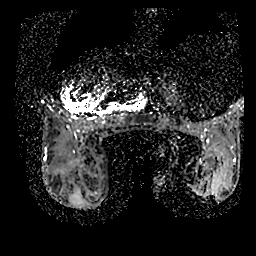}
        \caption{False positive prediction.}
        \label{fig:fp}
    \end{subfigure}
    \hfill
    \begin{subfigure}[b]{0.45\linewidth}
        \centering
        \includegraphics[width=\linewidth]{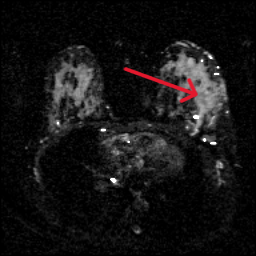}
        \caption{False negative prediction.}
        \label{fig:fn}
    \end{subfigure}
    \caption{Classification results: ADC slices incorrectly classified by the model.}
\end{figure}

Our model is able to correctly classify the majority of slices, performing especially well on cases with clearly defined tumors, such as the one shown on the right breast in Figure~\ref{fig:tp} (see arrow). The model also demonstrates good specificity. In Figure~\ref{fig:tn}, we show how the model correctly identified this slice as non-cancerous despite the presence of multiple dark regions in the ADC. However, certain challenging cases affect the classification performance. For instance, in Figure~\ref{fig:fp}, the model mistakenly classified the slice as "tumor present", misled by the low ADC sparse mass on the right-bottom side of the image. The slice in Figure~\ref{fig:fn} was incorrectly classified as non-cancerous but it was annotated as cancerous in the ground truth (see arrow).

\subsection{Tumor detection and localization}
Here, we study the object detection model's ability to localize tumors placing bounding boxes. Although detection does not provide detailed shape information, it is useful in clinical scenarios where quick localization is required. 

\begin{table}
    \centering
     \caption{Performance of the model in tumor detection and localization.}
    \begin{tabular}{|c|c|c|c|}
    \hline
         \textbf{Fold} & \textbf{Precision} & \textbf{Recall} & \textbf{AP}\\
    \hline
         1 & 0.917 & 0.767 & 0.840\\
         2 & 0.894 & 0.671 & 0.775\\
         3 & 0.864 & 0.634 & 0.722\\
         4 & 0.842 & 0.649 & 0.730\\
         5 & 0.878 & 0.681 & 0.724\\
         \textbf{Mean} & \textbf{0.878} & \textbf{0.680} & \textbf{0.758} \\
    \hline
    \end{tabular}
    \label{tab:detection}
\end{table}

\begin{table}
    \centering
     \caption{Performance in breast cancer tumor semantic segmentation.}
    \begin{tabular}{|c|c|}
    \hline
         \textbf{Model} & \textbf{DICE}   \\
    \hline
         \textbf{2D} & 0.699  \\
         \textbf{3D} & 0.696\\
    \hline
    \end{tabular}
    \label{tab:segmentation}
\end{table}

The model achieved a precision of 87.8\%, a recall of 68.0\% and an average precision of 75.8\%; see Table~\ref{tab:detection}. These values indicate a reasonable ability to localize suspicious regions in ADC images.

%When comparing detection with segmentation, it is important to consider that both tasks operate on different levels of granularity. Precision and recall can still be interpreted in both cases, but while segmentation evaluates overlap at the pixel level, detection considers it at the region level. Despite this, object detection shows higher values in overlap-related metrics, which suggests a stronger ability to point out tumor presence.

\begin{figure}
    \centering
    \begin{subfigure}[b]{0.45\linewidth}
        \centering
        \includegraphics[width=\linewidth]{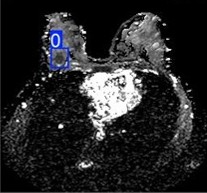}
        \caption{Annotated ground truth.}
        \label{fig:success1}
    \end{subfigure}
    \hfill
    \begin{subfigure}[b]{0.45\linewidth}
        \centering
        \includegraphics[width=\linewidth]{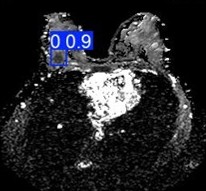}
        \caption{True positive prediction.}
        \label{fig:success2}
    \end{subfigure}
    \caption{Localization results: (a) The ground truth shows the tumor enclosed by a blue box and assigned the label 0. (b) The tumor is enclosed in a blue box labeled as class 0, with the confidence score (0.9) shown next to the label.}
    \label{fig:yolo_success}
\end{figure}

\begin{figure}
    \centering
    \begin{subfigure}[b]{0.45\linewidth}
        \centering
        \includegraphics[width=\linewidth]{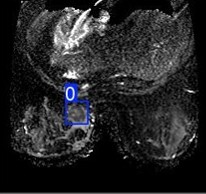}
        \caption{Annotated ground truth.}
        \label{fig:failure1}
    \end{subfigure}
    \hfill
    \begin{subfigure}[b]{0.45\linewidth}
        \centering
        \includegraphics[width=\linewidth]{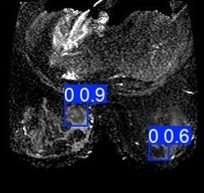}
        \caption{True and false positive predictions.}
        \label{fig:failure}
    \end{subfigure}
    \caption{Localization results: (a) The ground truth shows the tumor enclosed by a blue box and assigned the label 0. (b) The model prediction shows two boxes with label 0, one with confidence score 0.9 (correct) and one with confidence score 0.6 (incorrect).}
    \label{fig:yolo_failure}
\end{figure}

Figure~\ref{fig:yolo_success} shows a successful example of tumor identification. Panel (a) shows the ground truth, where the tumor is enclosed by a blue box and marked by the label 0. The right hand side shows the model's prediction. The model correctly identified the tumor location and assigned to it the 0 label and a confidence score of 0.9 (blue tag). The tumor is close to the rib cage and shows as a rounded lesion with restricted diffusion surrounded by tissue with less resistance to water diffusion. 

Figure~\ref{fig:yolo_failure} shows both a true positive and a false positive in the same slice. The ground truth image [panel (a)] shows a tumor in the left breast. However, as shown in panel (b), although the model correctly predicts the existing tumor, it also assigns a 0.6 confidence score to non-cancerous tissue on the right breast. A potential reason for the erroneous localization is that the box with confidence score 0.6 shows a low ADC signal, which is also common in tumors.

%the model can also produce errors, such as false positives, by classifying other low-intensity regions as tumors. These errors highlight the challenge of differentiating between pathological features and benign structures or artifacts that exhibit similar characteristics.

\subsection{Semantic segmentation of cancerous tumors}

Here, the goal is to generate pixel-wise masks of cancerous tumors. As displayed in Table~\ref{tab:segmentation}, 2D and 3D models showcase similar performance in terms of the DICE metric. The high inter-slice distance present in the dataset ($\sim$3 to 5 mm) likely greatly hinders the performance achieved by the 3D U-Net, as the similar DICE scores obtained by both the 3D and 2D nnU-Net models point to a struggle of the networks to learn the 3D structure of the tumor.

Nevertheless, all networks present subpar metrics, especially when compared with contrast-based imaging \cite{xia_early_2024}. The results clearly display the shortcomings of ADC for segmentation, especially considering that the nnU-Net models, which are known to achieve strong baseline performance, still fall short of the state-of-the-art performance reported for breast cancer segmentation.

% Although it is expected that 3D models should perform better due to the richer spatial context, the results indicate the opposite. The 2D model achieved better results across all metrics except recall (Table~\ref{tab:segmentation}), with a Dice score of 0.609 versus 0.463 in the 3D case.

%These results show a clear advantage for 2D segmentation over 3D. This could be explained by the anatomical nature of DW-MRI: the 3-5 mm slice thickness introduces large gaps in the volume, which make 3D connections between slices unreliable. 

%Nevertheless, the dataset presents some limitations. Despite bexºing labeled by experts and being used across various research publications \cite{newitt_testretest_2019} \cite{ partridge_diffusion-weighted_2018}, the complexity of the labeling process creates discrepancies, even between experts, resulting in subjective masks and bounding boxes that may hinder the learning on the models. Furthermore, the 4 mm gap between consecutive slices in the diffusion-weighted (DW) images posed a challenge for accurate volume reconstruction, necessitating the use of interpolation techniques to improve the continuity and spatial consistency of the reconstructed volume.

\section{Conclusion}\label{sec: conclusion}

This study demonstrates the potential of ADC maps derived from DW-MRI for breast cancer detection. Slice classification yielded encouraging results, with precision and accuracy levels that support the use of DW-MRI as an aid in screening applications for patients with dense breast tissue \cite{kim_highresolution_2024}. However, the relatively low recall should be carefully considered, as the risk of false negatives may limit its real-world applicability. Object detection also achieved good lesion localization performance, particularly given the inherent limitations of DW-MRI, including low spatial resolution and imaging artifacts \cite{schakel_evaluation_2018}. Nevertheless, object detection exhibited similarly limited recall, which may restrict its integration into screening workflows.

For segmentation, however, DW-MRI still presents significant challenges. The limited performance obtained highlights the difficulties of diffusion-based tumor segmentation, resulting in only moderate DICE scores. Furthermore, the large inter-slice spacing of the dataset leads to highly anisotropic volumes, reducing the continuity of anatomical information across slices. This characteristic likely contributes to the similar performance observed between the 2D and 3D models, suggesting that the available volumetric information is insufficient to fully exploit the advantages of 3D segmentation architectures.

Further research may explore the integration of advanced interpolation methods, such as the work presented by Zhang \emph{et al.} \cite{zhang_large-scale_2024}, applying DL techniques in the reconstruction of the 3D volumes. Other studies have achieved noteworthy results through the use of multimodal data, such as \cite {cai_diagnosis_2014, wang_breast_2024} with their combined use of DCE-MRI and DW-MRI. A DL model trained with both DW-MRIs and ADC maps as input could achieve increased results. Improvements in annotation quality could also optimize the metrics achieved \cite{ yu_robustness_2020}. As described by Ma \emph{et al.} \cite{ma_quantitative_2024} partial annotations by radiologist may significantly hinder the metrics achieved by the algorithms. These directions could help bridge the gap between the performance of DW-MRI and DCE-MRI, offering faster and contrast-free alternatives for breast cancer diagnosis.

\section*{Acknowledgements}
This research has been partially funded by the Council of Gijón through the University Institute of Industrial Technology of Asturias (IUTA) grants SV-25-GIJON-1-14, SV-25-GIJON-1-02, SV-24-GIJON-1-05, SV-24-GIJON-1-18, SV-24-GIJON-1-16, SV-23-GIJON-1-09, SV-22-GIJON-1-19, and SV-21-GIJON-1-19, and by Principado de Asturias, grant SV-PA-21-AYUD/2021/50994. 
GL acknowledges grant RYC2022-036010-I funded by MICIU/AEI/10.13039/501100011033 and ESF+. 

\section*{CRediT authorship contribution statement}
\textbf{Paula Puerta González:} Writing – review and editing, Writing – original draft, Visualization, Validation, Investigation, Data curation.  \textbf{Pablo García Marcos:} Writing – review and editing, Writing – original draft, Visualization, Validation, Investigation, Data curation. \textbf{Guillermo Lorenzo:} Writing – review and editing, Writing – original draft, Visualization, Validation, Methodology, Investigation, Conceptualization. \textbf{Hector Gomez:} Writing – review and editing, Writing – original draft, Visualization, Validation, Methodology, Investigation, Conceptualization. \textbf{Covadonga del Camino:} Writing – review and editing, Writing – original draft, Validation, Investigation, Conceptualization, Resources. \textbf{Víctor M. González:} Writing – review and editing, Writing – original draft, Visualization, Validation, Methodology, Investigation, Formal analysis, Data curation, Conceptualization, Project Administration, Resources, Funding acquisition. \textbf{Angel Rio-Alvarez:} Writing – review and editing, Writing – original draft, Visualization, Validation, Methodology, Investigation, Formal analysis, Data curation, Conceptualization, Project administration, Supervision. 

\section*{Ethics statement}
All procedures in this research were conducted in compliance with applicable laws and institutional guidelines.
This study leveraged publicly available, fully de-identified imaging data of breast cancer patients obtained from The Cancer Imaging Archive \cite{noauthor_acrin_nodate}, for which Institutional Review Board approval and informed consent were not required.
The authors confirm that no ethical complications were identified during the course of the study.
%All procedures in this research were conducted in compliance with applicable laws and institutional guidelines. The research did not require ethical approval as it was not conducted on human subjects or animals  . The authors confirm that no ethical complications were identified during the course of the study.

%\hl{guille, is this accurate?}\gledit{I have edited the statement to address a few issues and got o the point}

\section*{Declaration of competing interest}
The authors declare that they have no known competing financial interests or personal relationships that could have appeared to influence the work reported in this paper.

\section*{Data availability}
The data used in this study are publicly available at The Cancer Imaging Archive \cite{noauthor_acrin_nodate}.

\newpage

\bibliography{bibliography}

\end{document}